%% file: PaperForReview.tex
\documentclass[10pt,twocolumn,letterpaper]{article}
\usepackage{wacv}

\usepackage{amsmath}
\usepackage{amssymb}
\usepackage{array}
\usepackage{bbold}
\usepackage{booktabs}
\usepackage{caption}
\usepackage{courier}
\usepackage{float}
\usepackage{graphicx}
\usepackage{helvet}
\usepackage{inconsolata}
\usepackage{latexsym}
\usepackage{microtype}
\usepackage{multirow}
\usepackage{subcaption}
\usepackage{times}
\usepackage{url}
\usepackage{xcolor}

\usepackage[pagebackref,breaklinks,colorlinks]{hyperref}

\usepackage[capitalize]{cleveref}
\crefname{section}{Sec.}{Secs.}
\Crefname{section}{Section}{Sections}
\Crefname{table}{Table}{Tables}
\crefname{table}{Tab.}{Tabs.}

\def\wacvPaperID{1815}
\def\confName{WACV}
\def\confYear{2024}

\begin{document}

\title{A Diffusion-based Method for Multi-turn Compositional Image Generation}

\author{
Chao Wang \\
Toronto AI Lab \\
LG Electronics \\
\texttt{chao2.wang@lge.com}
}

\maketitle

\begin{abstract}
Multi-turn compositional image generation (M-CIG) is a challenging task that aims to iteratively manipulate a reference image given a modification text. While most of the existing methods for M-CIG are based on generative adversarial networks (GANs), recent advances in image generation have demonstrated the superiority of diffusion models over GANs. In this paper, we propose a diffusion-based method for M-CIG named conditional denoising diffusion with image compositional matching (CDD-ICM). We leverage CLIP as the backbone of image and text encoders, and incorporate a gated fusion mechanism, originally proposed for question answering, to compositionally fuse the reference image and the modification text at each turn of M-CIG. We introduce a conditioning scheme to generate the target image based on the fusion results. To prioritize the semantic quality of the generated target image, we learn an auxiliary image compositional match (ICM) objective, along with the conditional denoising diffusion (CDD) objective in a multi-task learning framework. Additionally, we also perform ICM guidance and classifier-free guidance to improve performance. Experimental results show that CDD-ICM achieves state-of-the-art results on two benchmark datasets for M-CIG, i.e., CoDraw and i-CLEVR.
\end{abstract}

\section{Introduction}
\begin{figure}[t]
\centering
\includegraphics[width=0.95\columnwidth,trim={0cm 2.5cm 11cm 0cm},clip]{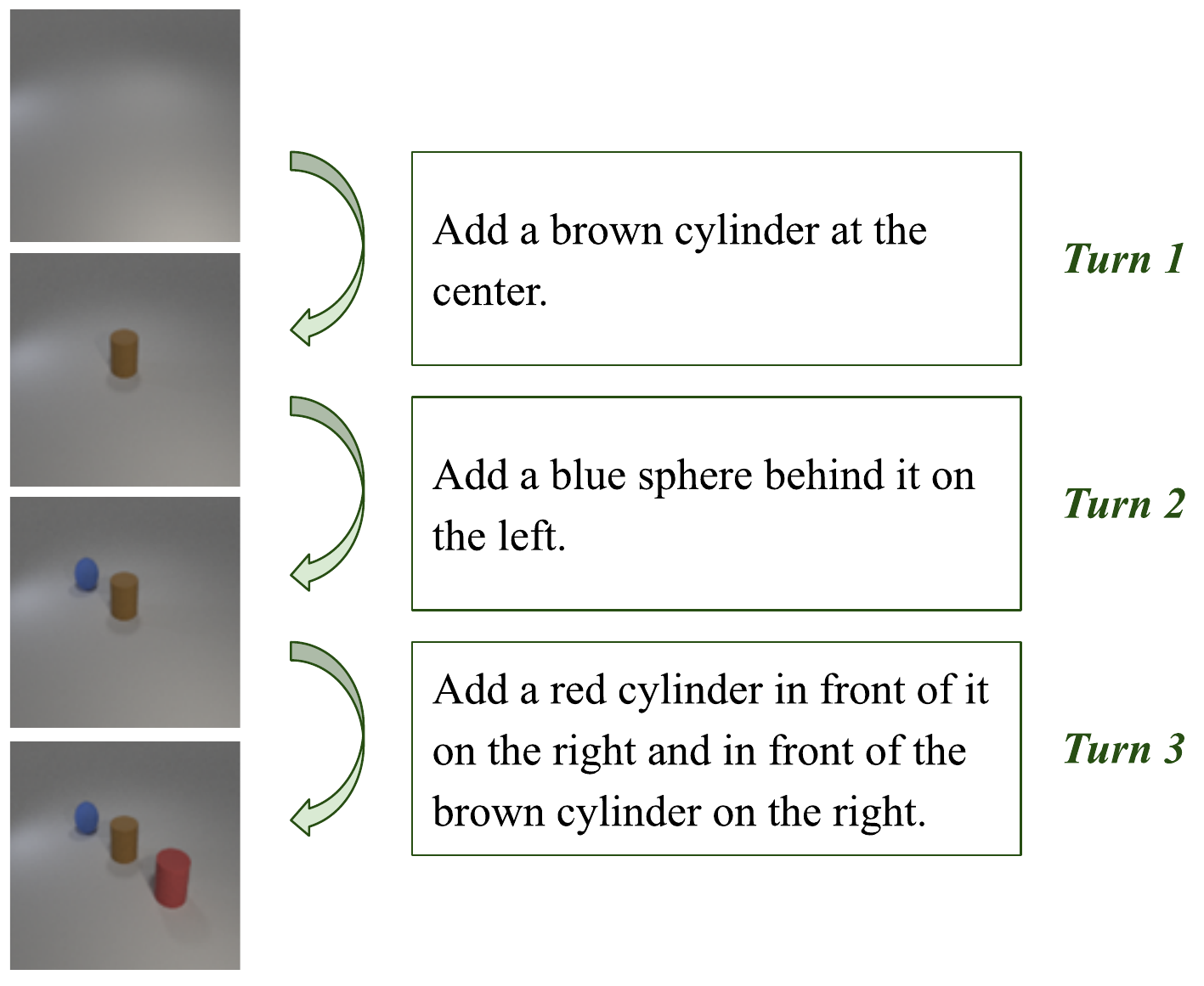}
\caption{A three-turn example of multi-turn compositional image generation (M-CIG).}
\label{f1}
\end{figure}
Image generation is a hot topic in computer vision, which has many applications in a wide range of areas, such as art, education, and entertainment. The generation of an image often needs to follow a text prompt. Additionally, sometimes the generation also needs to be based on an existing image rather than starting from scratch. Combining the above two requirements brings about \textit{compositional image generation (CIG)}, which is to generate a target image by changing a reference image according to a modification text. Addressing this cross-modal task is useful in computer-aided design (CAD), as it enables a computer system to generate images given verbal instructions from users.

In this paper, we focus on \textit{multi-turn compositional image generation (M-CIG)}, which is to perform CIG in an iterative manner. As shown in Figure~\ref{f1}, M-CIG can be described as a sequence of CIG turns, where the initial reference image is a background canvas, and the target image generated at each turn will be used as the reference image at the next turn. Compared with CIG, M-CIG is more challenging due to the iterative setting. Meanwhile, M-CIG is also more practical than CIG, as in the real world, a user usually needs to go through a series of incremental interactions with a computer system before achieving a final goal.

To the best of our knowledge, the existing methods for M-CIG \cite{el2019tell,fu-etal-2020-sscr,matsumori2021lattegan} are mostly based on generative adversarial networks (GANs) \cite{goodfellow2014generative}, which are currently the dominant family of techniques in image generation. According to some theoretical and empirical studies \cite{salimans2016improved,arjovsky2017towards,brock2017neural,miyato2018spectral,brock2019large}, although GANs can generate high-quality images, they are usually difficult to train, and the diversity of the generated images is also limited. Recently, diffusion models \cite{sohl2015deep,song2019generative,ho2020denoising,song2021denoising}, which are another family of generative modeling techniques, have gained great popularity in image generation. Compared with GANs, diffusion models are easier to train due to the straightforward definition of objectives, and can also generate more diverse images due to the explicit modeling of data distribution. As for the quality of the generated images, it has been demonstrated that diffusion models are comparable to or even better than GANs \cite{song2021scorebased,nichol2021improved,dhariwal2021diffusion,ho2022cascaded}. Therefore, we apply diffusion models to M-CIG.

Diffusion models rely on a denoising diffusion mechanism \cite{ho2020denoising} to generate images, which can be conditional so that only desired images are generated. Therefore, the key to addressing M-CIG using diffusion models is to learn conditional denoising diffusion (CDD), where the condition for generating the target image at each turn comes from the reference image and the modification text. However, this raises the following two problems:

\textbf{The lack of an appropriate conditioning scheme.} Although many conditioning schemes have been proposed for diffusion models, most of these works only deal with uni-modal cases, where the condition comes from either an image \cite{saharia2022palette,saharia2022image,rombach2022high} or a text \cite{nichol2022glide,ramesh2022hierarchical,saharia2022photorealistic}. The conditioning scheme proposed by \cite{kim2022diffusionclip} is aimed at a multi-modal case, where the condition comes from an image-text pair, but this work assumes that the text just describes the semantics of the image rather than the desired change to it. In a word, the above conditioning schemes cannot support the application of diffusion models to M-CIG.

\textbf{The concern about the semantic quality of the generated target image.} For the generated target image, we are not only concerned with its visual quality, but also its semantic quality, which refers to whether it contains the desired objects and whether the contained objects constitute the desired topology. Actually, we believe that the semantic quality deserves more concern than the visual quality. The reason is two-fold. On the one hand, a high semantic quality implies a high visual quality, but the reverse is not true. On the other hand, due to the iterative nature of M-CIG, semantic mistakes are likely to accumulate from turn to turn, which may corrupt the rear turns.

To solve these problems, we propose a diffusion-based method for M-CIG named \textit{conditional denoising diffusion with image compositional matching (CDD-ICM)}, which features a novel conditioning scheme equipped with a multi-task learning framework. Specifically, we use CLIP \cite{radford2021learning} as the backbone to encode images and texts. On this basis, we borrow a gated fusion mechanism from a question answering (QA) method \cite{wang-etal-2018-multi-granularity} to perform compositional fusion between the reference image and the modification text at each turn of M-CIG, and use the result as the condition of the denoising diffusion mechanism to generate the target image. To guarantee the semantic quality of the generated target image, we learn image compositional matching (ICM) as an auxiliary objective of CDD to explicitly enhance the conditon, where the compositional fusion result is aligned with the representation of the target image through contrastive learning. Moreover, we also perform ICM guidance and classifier-free guidance \cite{ho2022classifier} to boost performance. Experimental results show that CDD-ICM achieves state-of-the-art (SOTA) performance on two benchmark datasets for M-CIG, namely CoDraw \cite{kim-etal-2019-codraw} and i-CLEVR \cite{el2019tell}.

The contribution of this paper is three-fold. First, we creatively apply diffusion models to M-CIG, where a novel conditioning scheme is developed to handle a compositional image-text pair, integrating the denoising diffusion mechanism with CLIP and a gated fusion mechanism. Second, to prioritize the semantic quality of the generated target image, we establish a multi-task learning framework for the conditioning scheme, where ICM serves as an auxiliary objective of CDD to explicitly enhance the condition. Third, our diffusion-based method outperforms the existing GAN-based methods on two M-CIG benchmark datasets.

\section{Method}
\begin{figure}[t]
\centering
\includegraphics[width=0.95\columnwidth,trim={5cm 1.5cm 9.3cm 1.3cm},clip]{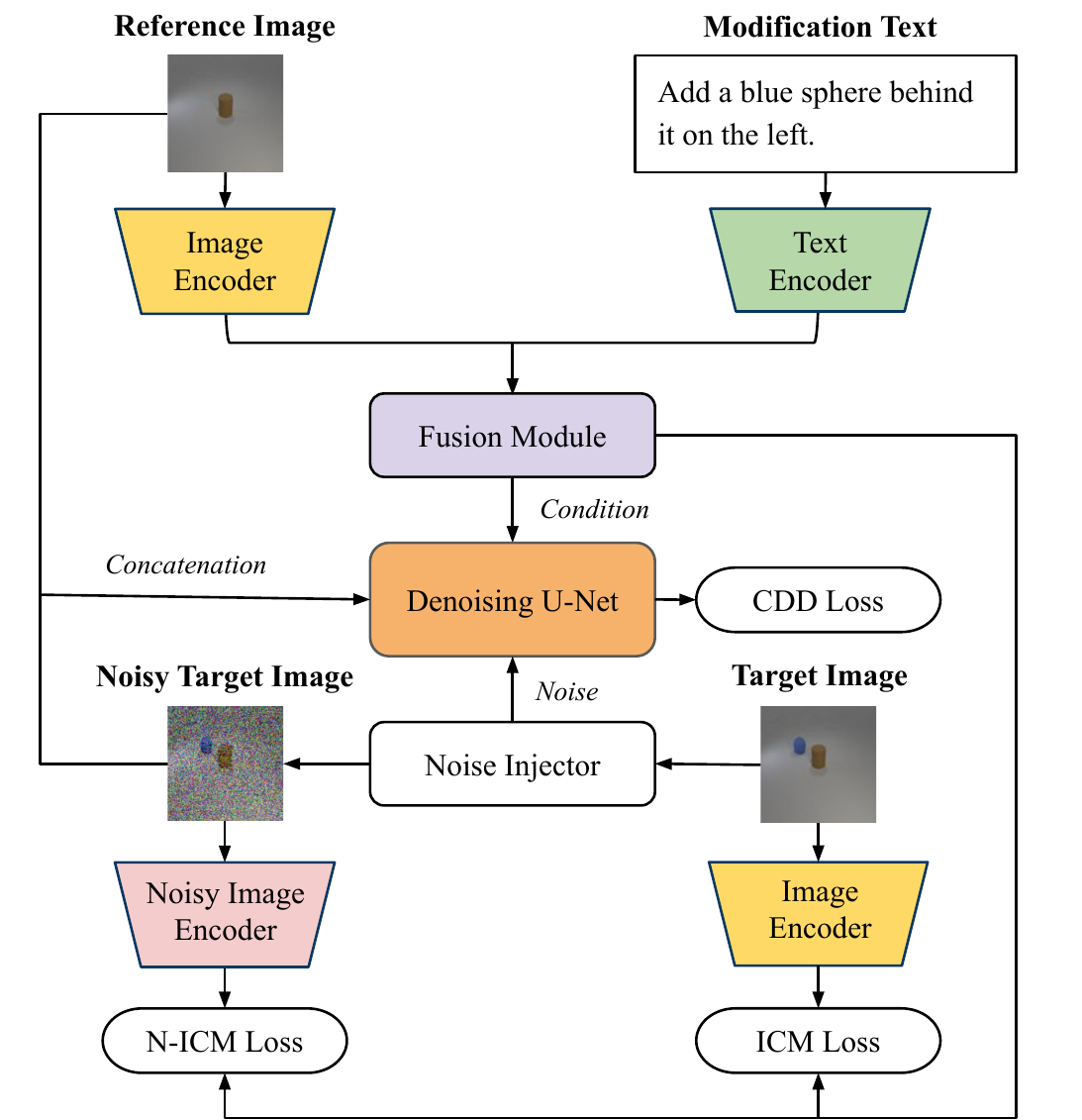}
\caption{An overview of CDD-ICM. Colored components are trainable, and those of the same color share their trainable parameters.}
\label{f2}
\end{figure}
In this section, we elaborate CDD-ICM, which is our diffusion-based method for M-CIG. We begin by providing a task formulation of M-CIG, which is followed by a detailed introduction to the design of CDD-ICM.

\subsection{Task Formulation}
Given an initial reference image $a^{(1)}$, which is a background canvas, and a sequence of $k$ modification texts $\{ m^{(1)}, \ldots, m^{(k)} \}$, which describe the desired changes to be successively made to $a^{(1)}$, M-CIG is a $k$-turn iterative process, where at each turn $i \in \{ 1, \ldots, k \}$, it is required to generate a target image $z^{(i)}$ by changing the current reference image $a^{(i)}$ according to $m^{(i)}$, and if $i < k$, $z^{(i)}$ will be used as the next reference image $a^{(i + 1)}$.

\subsection{Encoding}
Considering the cross-modal nature of M-CIG, we map images and texts into a joint representation space through encoding. As shown in Figure~\ref{f2}, we include an image encoder and a text encoder in CDD-ICM, where the former is used to encode reference images and target images, and the latter is used to encode modification texts. Since M-CIG is a vision-and-language (V\&L) task, we take advantage of large-scale V\&L pre-training by using CLIP, which is pre-trained on 400M image-text pairs, as the backbone of both encoders. Specifically, we use the vision part of CLIP, which is a vision transformer (ViT) \cite{dosovitskiy2021an}, as the backbone of the image encoder, and use the language part of CLIP, which is a GPT-like \cite{radford2018improving} auto-regressive language model, as the backbone of the text encoder. In each encoder, we append a linear projection layer after the backbone, which finally yields the representations. For both linear projection layers, we set their output dimensionality to the same value $d$, which is a hyper-parameter denoting the dimensionality of the joint representation space.

Additionally, as shown in Figure~\ref{f2}, we also include a noisy image encoder in CDD-ICM, which is used to encode the noisy target images obtained in the denoising diffusion mechanism. It has the same structure as the image encoder, but holds different trainable parameters.

\subsection{Compositional Fusion}
At each turn of M-CIG, to extract clues for the generation of the target image, we perform compositional fusion between the reference image and the modification text. As shown in Figure~\ref{f2}, we include a fusion module in CDD-ICM, which fuses the representation of the reference image with that of the modification text. Actually, we can interpret each turn of M-CIG from the perspective of QA. Specifically, we can regard the reference image as a context, the modification text as a relevant question, and the target image as the corresponding answer. In this way, to implement the fusion module, we borrow the following gated fusion mechanism from a QA method \cite{wang-etal-2018-multi-granularity}:
\begin{equation}
\small
\begin{split}
& \mathrm{f} ( u, v ) = g \odot h + ( 1 - g ) \odot u \\
& g = \mathrm{sigmoid} ( W_g [ u; v; u \odot v; u - v ] + b_g ) \\
& h = \mathrm{gelu} ( W_h [ u; v; u \odot v; u - v ] + b_h )
\end{split}
\label{e1}
\end{equation}
where $W_g$ and $W_h$ are trainable weight matrices, $b_g$ and $b_h$ are trainable bias vectors, $\odot$ denotes element-wise multiplication, and $[ ; ]$ denotes vector concatenation. In the fusion module, we set $u$ to the representation of the reference image, set $v$ to that of the modification text, and thereby obtain $\mathrm{f} ( u, v )$ as the compositional fusion result.

\subsection{Conditional Denoising Diffusion}
To generate the target image at each turn of M-CIG, we learn conditional denoising diffusion (CDD), which is to perform the generation using a denoising diffusion mechanism conditioned on the compositional fusion result between the reference image and the modification text. As proposed by \cite{ho2020denoising}, the denoising diffusion mechanism consists of a forward diffusion process, which gradually injects noises to the target image, and a reverse denoising process, which gradually erases the injected noises.

With the target image denoted as $x_0 \sim q ( x_0 )$, the forward diffusion process is a pre-defined Markov chain of $T$ time steps $\langle 1, \ldots, T \rangle$, where the state transfers from $x_0$ all the way to $x_T$. Specifically, at each time step $t$, we obtain $x_t$ by injecting a Gaussian noise $\epsilon_t \sim N ( 0, \mathit{I} )$ to $x_{t-1}$:
\begin{equation}
\small
\begin{split}
& q ( x_t | x_{t - 1} ) = N ( \sqrt{\alpha_t} x_{t - 1}, ( 1 - \alpha_t ) \mathit{I} ) \\
\Leftrightarrow \quad & x_t = \sqrt{\alpha_t} x_{t - 1} + \sqrt{1 - \alpha_t} \epsilon_t
\end{split}
\label{e2}
\end{equation}
where $\alpha_t$ is a hyper-parameter used to control the noise scale. It is easy to derive that with $\prod_{i = 1}^t \alpha_i$ denoted as $\bar{\alpha}_t$, we can actually obtain $x_t$ by directly injecting $\epsilon_t$ to $x_0$:
\begin{equation}
\small
\begin{split}
& q ( x_t | x_0 ) = N ( \sqrt{\bar{\alpha}_t} x_0, ( 1 - \bar{\alpha}_t ) \mathit{I} ) \\
\Leftrightarrow \quad & x_t = \sqrt{\bar{\alpha}_t} x_0 + \sqrt{1 - \bar{\alpha}_t} \epsilon_t
\end{split}
\label{e3}
\end{equation}
As shown in Figure~\ref{f2}, we include a noise injector in CDD-ICM, which executes Equation~\ref{e3} at an arbitrary time step $t$ to obtain $x_t$ as a noisy target image. To implement the noise injector, we set each $\alpha_t \in \{ \alpha_1, \ldots, \alpha_T \}$ in the following way proposed by \cite{nichol2021improved}:
\begin{equation}
\small
\begin{split}
& \alpha_t = \frac{s ( t )}{s ( t - 1 )} \\
& s ( t ) = \mathrm{cos}^2 ( \frac{t / T + 0.008}{1.008} \cdot \frac{\pi}{2} )
\end{split}
\label{e4}
\end{equation}

With the compositional fusion result between the reference image and the modification text denoted as $c$, the reverse denoising process is a parameterized Markov chain of $T$ time steps $\langle T, \ldots, 1 \rangle$, where the state transfers from $x_T$ all the way back to $x_0$ conditioned on $c$. Specifically, at each time step $t$, given $x_t$ and the condition $c$, we obtain $x_{t - 1}$ using a neural network $\theta$:
\begin{equation}
\small
\begin{split}
& p_\theta ( x_{t - 1} | x_t, c ) = N \big( \mu_\theta ( x_t, t, c ), \Sigma_\theta ( x_t, t, c ) \big) \\
\Leftrightarrow \quad & x_{t - 1} = \mu_\theta ( x_t, t, c ) + \Sigma_\theta^{\frac{1}{2}} ( x_t, t, c ) \xi_t
\end{split}
\label{e5}
\end{equation}
where $\xi_t \sim N ( 0, \mathit{I} )$. According to \cite{ho2020denoising}, we can learn $\theta$ by minimizing the following variational lower-bound (VLB) loss:
\begin{equation}
\small
\begin{split}
& L^{vlb} = E_{x_0 \sim q ( x_0 ), t \sim U \{ 1, T \}} [ L^{vlb}_t ] \\
& L^{vlb}_t =
\begin{cases}
-\mathrm{log} \ p_\theta ( x_{t - 1} | x_t, c ), & \text{if} \ t = 1 \\
D_{KL} \big( q( x_{t - 1} | x_t, x_0 ) || p_\theta ( x_{t - 1} | x_t, c ) \big), & \text{else}
\end{cases}
\end{split}
\label{e6}
\end{equation}
Using Bayes' theorem, it can be derived that with $1 - \alpha_t$ denoted as $\beta_t$, $q( x_{t - 1} | x_t, x_0 )$ is the following Gaussian distribution:
\begin{equation}
\small
\begin{split}
& q( x_{t - 1} | x_t, x_0 ) = N \big( \tilde{\mu} ( x_t, x_0 ), \tilde{\beta}_t \mathit{I} \big) \\
& \tilde{\mu} ( x_t, x_0 ) = \frac{\sqrt{\bar{\alpha}_{t - 1} \beta_t}}{1 - \bar{\alpha}_t} x_0 + \frac{\sqrt{\alpha_t} ( 1 - \bar{\alpha}_{t - 1} )}{1 - \bar{\alpha}_t} x_t \\
& \tilde{\beta}_t = \frac{1 - \bar{\alpha}_{t - 1}}{1 - \bar{\alpha}_t} \beta_t
\end{split}
\label{e7}
\end{equation}
Based on Equation~\ref{e3} and Equation~\ref{e7}, we parameterize $\mu_\theta ( x_t, t, c )$ in the following way proposed by \cite{ho2020denoising}:
\begin{equation}
\small
\mu_\theta ( x_t, t, c ) = \frac{1}{\sqrt{\alpha_t}} \big( x_t - \frac{\beta_t}{\sqrt{1 - \bar{\alpha}_t}} \epsilon_\theta ( x_t, t, c ) \big)
\label{e8}
\end{equation}
where $\epsilon_\theta ( x_t, t, c )$ is a prediction to $\epsilon_t$. Besides, we also parameterize $\Sigma_\theta ( x_t, t, c )$ in the following way proposed by \cite{nichol2021improved}:
\begin{equation}
\small
\Sigma_\theta ( x_t, t, c ) = \mathrm{exp} \big( \mathrm{log} \frac{\beta_t}{\tilde{\beta}_t} \rho_\theta ( x_t, t, c ) + \mathrm{log} \tilde{\beta}_t \big)
\label{e9}
\end{equation}
where $\rho_\theta ( x_t, t, c )$ is a fraction used to interpolate between $\mathrm{log} \beta_t$ and $\mathrm{log} \tilde{\beta}_t$. As shown in Figure~\ref{f2}, we include a denoising U-Net in CDD-ICM, which performs the above parameterizations and thus can be seen as $\theta$. To implement the denoising U-Net, we make three changes to the U-Net structure used by \cite{dhariwal2021diffusion}. First, we replace the class embedding with the condition $c$. Second, we concatenate $x_t$ with the reference image along the channel dimension, and thereby use the result as the input. Third, we concatenate the patch representations of the reference image with the token representations of the modification text, and thereby use the result to augment each attention layer as suggested by \cite{nichol2022glide}. The output of the denoising U-Net is divided into two parts along the channel dimension, which are separately used as $\epsilon_\theta ( x_t, t, c )$ and $\rho_\theta ( x_t, t, c )$.

According to \cite{ho2020denoising}, to learn the above reverse denoising process, instead of minimizing $L^{vlb}$, we can actually minimize the following mean squared error (MSE) loss:
\begin{equation}
\small
L^{mse} = E_{x_0 \sim q ( x_0 ), t \sim U \{ 1, T \}} [ || \epsilon_t - \epsilon_\theta ( x_t, t, c ) ||^2 ]
\label{e10}
\end{equation}
Compared with $L^{vlb}$, $L^{mse}$ is not only simpler but also more effective. However, minimizing $L^{mse}$ cannot bring any learning signal to $\Sigma_\theta ( x_t, t, c )$. To benefit from learning $\Sigma_\theta ( x_t, t, c )$, we combine $L^{mse}$ with $L^{vlb}$ as suggested by \cite{nichol2021improved}. Specifically, we minimize a CDD loss $L^{cdd}$, which is calculated as follows:
\begin{equation}
\small
L^{cdd} = L^{mse} + \gamma L^{vlb}
\label{e11}
\end{equation}
where $\gamma$ is a hyper-parameter used to control the weight of $L^{vlb}$. Additionally, we also stop the gradients of $L^{vlb}$ from flowing to $\epsilon_\theta ( x_t, t, c )$.

\subsection{Image Compositional Matching}
At each turn of M-CIG, the semantic quality of the generated target image depends on the condition of the denoising diffusion mechanism, which is the compositional fusion result between the reference image and the modification text. Although learning CDD ensures that the condition is learned, this effect is implicit. To explicitly enhance the condition so that it embodies more clues about the target image, we learn image compositional matching (ICM) as an auxiliary objective of CDD, which is to align the compositional fusion result with the representation of the target image.

To learn ICM, we adopt the InfoNCE loss \cite{oord2018representation} used in the contrastive pre-training of CLIP, and apply it to individual turns constituting M-CIG samples. Specifically, given a mini-batch of $n$ (reference image, modification text, target image) triples $\{ ( a_1, m_1, z_1 ), \ldots, ( a_n, m_n, z_n ) \}$, each of which denotes an individual turn picked from an M-CIG sample, we treat them as positive samples, and generate $n^2 - n$ negative samples by replacing the target image $z_i$ in each positive sample $( a_i, m_i, z_i )$ separately with the other $n - 1$ target images $\{ z_1, \ldots, z_n \} - \{ z_i \}$. For each of the positive and negative samples, suppose that we have already obtained the compositional fusion result between the reference image and the modification text and the representation of the target image, then we calculate the cosine similarity between them. As a result, we construct a similarity matrix $S \in \mathbb{R}^{n \times n}$, where the element at the $i$-th row and the $j$-th column corresponds to the sample $( a_i, m_i, z_j )$. It is easy to see that the diagonal elements in $S$ correspond to the positive samples, while the other elements correspond to the negative samples. Based on $S$, we minimize an ICM loss $L^{icm}$, which is calculated as follows:
\begin{equation}
\small
\begin{split}
L^{icm} = & \frac{1}{n} \mathrm{tr} \Big( \mathrm{-log} \big( \mathrm{softmax} ( \frac{S}{\tau} ) \big) \Big) + \\
& \frac{1}{n} \mathrm{tr} \Big( \mathrm{-log} \big( \mathrm{softmax} ( \frac{S^{\top}}{\tau} ) \big) \Big)
\end{split}
\label{e12}
\end{equation}
where $\tau$ is a trainable temperature scalar, $\mathrm{tr} ( \cdot )$ denotes calculating matrix trace, and $\mathrm{softmax} ( \cdot )$ is calculated along the row dimension. In this way, the compositional fusion result between a (reference image, modification text) pair will be close to the representation of the real target image, while apart from those of the fake ones.

Besides, to enable ICM guidance, which will be introduced later, we also learn noise-aware image compositional matching (N-ICM). Specifically, we replace the above target images with their noisy variants, which are obtained using the noise injector, and encode these noisy target images using the noisy image encoder. On this basis, we minimize an N-ICM loss $L^{n-icm}$, which is calculated in the same way as we calculate $L^{icm}$.

\subsection{Training and Inference}
For the training, we disassemble M-CIG samples into individual turns and thereby apply teacher forcing. On this basis, we divide the training into three stages, where in each stage, we minimize a different loss through mini-batch gradient descent to update the corresponding trainable components of CDD-ICM. Specifically, in the first stage, we minimize $L^{icm}$ to update the image encoder, the text encoder, and the fusion module. In the second stage, we minimize the following joint loss, which is a combination of $L^{cdd}$ and $L^{icm}$, to update the image encoder, the text encoder, the fusion module, and the denoising U-Net:
\begin{equation}
\small
L^{joint} = L^{cdd} + \delta L^{icm}
\label{e13}
\end{equation}
where $\delta$ is a hyper-parameter used to control the weight of $L^{icm}$. In the third stage, we freeze the image encoder, the text encoder, and the fusion module, and minimize $L^{n-icm}$ to update the noisy image encoder. To effectively fine-tune CLIP, we set the backbone learning rate as a product of the global learning rate and a backbone activity ratio $\eta$, which is a hyper-parameter. From the perspective of transfer learning, $\eta$ controls the trade-off between the knowledge transferred from CLIP and that embodied in the training data.

For the inference, at each turn of M-CIG, we use the image encoder to encode the reference image, use the text encoder to encode the modification text, use the fusion module to perform compositional fusion based on the encoding results, and use the denoising U-Net to iteratively execute Equation~\ref{e5} from the time step $T$ until the time step $1$, where the condition $c$ is set to the compositional fusion result. From Equation~\ref{e3} and Equation~\ref{e4}, it can be derived that if $T$ is large enough, then $q( x_T ) \approx N ( 0, I )$, thus we sample $x_T$ from $N ( 0, I )$ at the time step $T$. To accelerate the inference, we traverse only a part of the time steps, which are uniformly distributed among all of them, and make this process deterministic as suggested by \cite{song2020denoising}. Finally, we obtain $x_0$ as the generated target image at the time step $1$.

Moreover, we also perform ICM guidance and classifier-free guidance to boost performance. Our ICM guidance is similar to the CLIP guidance of \cite{nichol2022glide}. Specifically, suppose that we have minimized $L^{n-icm}$ in the training, then in the inference, instead of using $\mu_\theta ( x_t, t, c )$ in Equation~\ref{e5}, we use $\hat{\mu}_\theta ( x_t, t, c )$, which is obtained by perturbing $\mu_\theta ( x_t, t, c )$ using the gradient of $L^{n-icm}$ with respect to $x_t$:
\begin{equation}
\small
\hat{\mu}_\theta ( x_t, t, c ) = \mu_\theta ( x_t, t, c ) + \psi \Sigma_\theta ( x_t, t, c ) \nabla_{x_t} L^{n-icm}
\label{e15}
\end{equation}
where $\psi$ is a hyper-parameter used to control the perturbation scale. Our classifier-free guidance is similar to that of \cite{nichol2022glide}. Specifically, in the training, when calculating $\epsilon_\theta ( x_t, t, c )$ in Equation~\ref{e10}, we set the condition $c$ to $\vec{0}$ with a probability of $\lambda$, which is a hyper-parameter. On this basis, in the inference, instead of using $\epsilon_\theta ( x_t, t, c )$ in Equation~\ref{e8}, we use $\hat{\epsilon}_\theta ( x_t, t, c )$, which is obtained by perturbing $\epsilon_\theta ( x_t, t, c )$ using $\epsilon_\theta ( x_t, t, \vec{0} )$:
\begin{equation}
\small
\hat{\epsilon}_\theta ( x_t, t, c ) = \phi \epsilon_\theta ( x_t, t, c ) + ( 1 - \phi ) \epsilon_\theta ( x_t, t, \vec{0} )
\label{e14}
\end{equation}
where $\phi$ is a hyper-parameter used to control the perturbation scale.

\section{Related Works}
\subsection{Image Manipulation}
The goal of image manipulation is to modify specific attributes of an image while avoiding unintended changes or generating a completely new image. Existing works can be split into two main categories: image-to-image translation and text-conditioned image manipulation.

\textbf{Image-to-Image Translation.} The image-to-image translation aims to generate an output image only conditioning on an input image, {\em i.e.}, uni-modal condition. Image inpainting and image super-resolution are two typical image-to-image translation tasks.
In recent years, deep learning has achieved great success in image inpainting. Context Encoders~\cite{pathak2016context} first explores to utilize conditional GANs. 
Multiple variants~\cite{wang2018image,yan2018shift,zeng2019learning,liu2020rethinking} of U-Net~\cite{ronneberger2015u} have been proposed for image inpainting. Some works explore multi-stage generation by taking object edges~\cite{nazeri2019edgeconnect}, structures~\cite{ren2019structureflow}, or semantic segmentation maps~\cite{song2018spg} as intermediate clues. 
In terms of super-resolution, most early works are regression-based and trained with MSE loss~\cite{dong2014learning,dong2015image,wang2015deep,kim2016accurate}. 
Auto-regressive models~\cite{dahl2017pixel,parmar2018image} and GAN-based methods~\cite{karras2017progressive,ledig2017photo,wang2018esrgan,menon2020pulse,lugmayr2020srflow} have also shown high quality results.

\textbf{Text-Conditioned Image Manipulation.} The text-conditioned image manipulation targets generating an output image conditioned on both the input image and text, {\em i.e.}, multi-modal condition. 
The input text can be a caption-like description of the target image, and the editing is usually single-turn. 
TAGAN~\cite{nam2018text} employs word-level local discriminators to preserve text-irrelevant content. ManiGAN~\cite{li2020manigan} first selects image regions and then correlates the regions with semantic words.
DiffusionCLIP~\cite{kim2022diffusionclip} is a robust framework that utilizes the pre-trained diffusion models and CLIP loss for image manipulation.

The input text can also be user-provided text instructions that describe desired modifications, such as adding, changing, or removing the objects in images. Generating an image based on provided instructions and an input image is dubbed as the compositional image generation (CIG) task in this paper. 
\cite{Ak2020LearningCR} achieve great performance on the benchmarks CSS~\cite{vo2019composing} and Fashion Synthesis~\cite{zhu2017your} by designing an improved image \& text composition layer and a multi-modal similarity module. 
\cite{zhang2021text} propose a GAN-based method to locally modify image features and show remarkable results on both CSS and Abstract Scene~\cite{zitnick2013bringing}. 
Afterward, the M-CIG task presents a more challenging setting compared to the above single-turn CIG task. \cite{el2019tell} first propose the M-CIG task known as Generative Neural Visual Artist (GeNeVA) task, which requires iteratively generating an image according to ongoing linguistic input. \cite{fu2020sscr} introduce the self-supervised counterfactual reasoning (SSCR) framework to tackle the data scarcity problem. LatteGAN~\cite{matsumori2021lattegan} improves desired object generation by introducing a Latte module and a text-conditioned U-Net discriminator. 
Our research work targets the M-CIG task, following~\cite{el2019tell}, we conducted experiments on CoDraw~\cite{kim2019codraw} and i-CLEVR~\cite{el2019tell}. 



\subsection{Diffusion Models}
Diffusion models (DMs)~\cite{sohl2015deep}, which formulate the data sampling process as an iterative denoising procedure, are closely related to a large family of methods for learning generative models as transition operators of Markov chains~\cite{bengio2014deep,sohl2015deep,salimans2015markov,song2017nice,levy2018generalizing}. 
Many research works concentrate on improving the diffusion process of DMs.
\cite{song2019generative} propose to estimate the gradients of data distribution via score matching and produce samples via Langevin dynamics.
Denoising diffusion probabilistic models (DDPMs)~\cite{ho2020denoising}, which optimize a variational lower bound to the log-likelihood, can achieve comparable sample quality as GANs~\cite{brock2018large,karras2019style}. 
Denoising diffusion implicit models (DDIMs)~\cite{song2021denoising} speed up the sampling process while enabling near-perfect inversion~\cite{dhariwal2021diffusion}.
The improved DDPM~\cite{nichol2021improved} introduces several modifications to achieve competitive likelihoods without sacrificing sample quality.
The latent diffusion~\cite{rombach2022high} model is applied in latent space instead of pixel space to enable an efficient diffusion process.
Although GANs have achieved plausible results in image synthesis, they are usually difficult to train and tend to limit the diversity of the generated images~\cite{shmelkov2018good,wang2021cycle}.
DMs are more stable during training and demonstrate comparable or even better performance for image synthesis~\cite{dhariwal2021diffusion,song2021scorebased,ho2022cascaded}.


Motivated by the progress in developing DMs, some research works explore text or image conditional diffusion mechanisms. 
While certain diffusion models solely utilize an input image for conditioning, such as PALETTE~\cite{saharia2022palette} and SR3~\cite{saharia2022image}, those that condition on both an input image and text are more pertinent to our work. 
GLIDE~\cite{nichol2022glide} is a text-guided diffusion model, where classifier-free guidance yields higher-quality images than CLIP guidance.
DALL-E 2~\cite{ramesh2022hierarchical} initially generates a CLIP~\cite{radford2021learning} image embedding given a text caption and then generates an image conditioned on the image embedding.
ImageGen~\cite{saharia2022photorealistic} exhibits a deep level of language understanding which enables high-fidelity image generation.
Stable Diffusion~\cite{rombach2022high} is an efficient latent diffusion model and has achieved superior image synthesis performance.
\cite{liu2022compositional} compose pre-trained text-guided diffusion models to improve structured generalization for image generation.
However, in the context of this paper, it should be noted that the textual input in the aforementioned models refers to a caption-like description, rather than iterative instructions on image manipulations. Designing diffusion-based methods for the M-CIG task remains a relatively underexplored area, presenting challenges in iteratively modifying images with instructions and conditioning the denoising diffusion mechanism on multi-modalities.
To address this gap in the literature, we propose a diffusion-based approach coupled with auxiliary ICM objectives to enhance the visual and semantic fidelity of generated images in the M-CIG task.

\section{Experiments}
\begin{table*}[ht!]
\small
\centering
\begin{tabular}{@{}ccccccccc@{}}
\toprule
\multirow{2}{*}{\textbf{Method}} & \multicolumn{4}{c}{\textbf{CoDraw}} & \multicolumn{4}{c}{\textbf{i-CLEVR}} \\
\cmidrule(l){2-5} \cmidrule(l){6-9} & \textbf{Precision} & \textbf{Recall} & \textbf{F1} & \textbf{RSIM} & \textbf{Precision} & \textbf{Recall} & \textbf{F1} & \textbf{RSIM} \\
\midrule
GeNeVA-GAN \cite{el2019tell} & 66.64 & 52.66 & 58.83 & 35.41 & 92.39 & 84.72 & 88.39 & 74.02 \\
SSCR \cite{fu-etal-2020-sscr} & 58.17 & 56.61 & 57.38 & 39.11 & 73.75 & 46.39 & 56.96 & 34.54 \\
TIRG \cite{kenan2020learning} & 76.56 & 73.40 & 72.40 & 46.64 & 94.30 & 92.96 & 93.71 & 77.55 \\
LatteGAN \cite{matsumori2021lattegan} & 81.50 & 78.37 & 77.51 & 54.16 & 97.72 & 96.93 & 97.26 & 83.21 \\
\midrule
CDD-ICM (ours) & \textbf{90.61} & \textbf{87.55} & \textbf{89.05} & \textbf{57.39} & \textbf{99.99} & \textbf{99.94} & \textbf{99.96} & \textbf{85.66} \\
\bottomrule
\end{tabular}
\caption{Performance comparison on CoDraw and i-CLEVR.}
\label{t1}
\end{table*}

\subsection{Datasets}
To verify the effectiveness of CDD-ICM, we conduct experiments on the following two M-CIG benchmark datasets:

\textbf{CoDraw.} CoDraw contains $8$K M-CIG samples for training, $1$K for validation, and $1$K for test. The number of turns per M-CIG sample varies between $1$ and $14$ with an average of $4.25$. The reference images and the target images contain $58$ classes of clip-art-style objects, such as boys, girls, and trees. The modification texts are conversations between a teller and a drawer.

\textbf{i-CLEVER.} i-CLEVER contains $6$K M-CIG samples for training, $2$K for validation, and $2$K for test. The number of turns per M-CIG sample is always $5$. The reference images and the target images contain $24$ classes of colored geometric objects, such as red spheres, yellow cubes, and blue cylinders. The modification texts are sentences indicating the addition of a new object.

Both datasets adopt four metrics to evaluate the semantic quality of the generated target images: precision, recall, F1 score, and relational similarity (RSIM). Specifically, each dataset comes with an object detector, which is trained on the dataset by \cite{el2019tell}. In the evaluation, the object detector is applied to both the generated target images and the ground-truth target images. On this basis, precision, recall, and F1 score are calculated for each turn of M-CIG by comparing the object presence in the generated target image with that in the ground-truth target image:
\begin{equation}
\small
\begin{split}
& \mathrm{precision} = \frac{| O_{gen} \cap O_{gt} |}{| O_{gen} |} \\
& \mathrm{recall} = \frac{| O_{gen} \cap O_{gt} |}{| O_{gt} |} \\
& \mathrm{F1} = 2 \cdot \frac{\mathrm{precision} \cdot \mathrm{recall}}{\mathrm{precision} + \mathrm{recall}}
\end{split}
\label{e16}
\end{equation}
where $O_{gen}$ and $O_{gt}$ denote the objects in the generated target image and the ground-truth target image, respectively. RSIM is calculated on the last turn of M-CIG by comparing the object topology in the generated target image with that in the ground-truth target image:
\begin{equation}
\small
\begin{split}
& \mathrm{RSIM} = \mathrm{recall} \cdot \frac{| E_{gen} \cap E_{gt} |}{| E_{gt} |}
\end{split}
\label{e17}
\end{equation}
where $E_{gen}$ and $E_{gt}$ denote the edges interconnecting $O_{gen} \cap O_{gt}$ in the generated target image and the ground-truth target image, respectively.

\subsection{Implementation Details}
We use PyTorch \cite{paszke2019pytorch} to implement CDD-ICM, and use HuggingFace's Transformers \cite{Wolf2019HuggingFacesTS} to load CLIP. In the three CLIP-based encoders, we adopt the basic version of CLIP (\textit{i.e.} CLIP-ViT-B/32) as the backbone, set the backbone activity ratio $\eta$ to $0.001$, and set the output dimensionality $d$ of all the linear projection layers to $512$. In the denoising diffusion mechanism, we use $1000$ time steps for the training (\textit{i.e.} $T = 1000$), and use the $250$ time steps uniformly distributed among them for the inference. For ICM guidance, we set $\psi$ to $2$. For classifier-free guidance, we set the value of $\lambda$ to $0.2$ and $\phi$ to $3$. To calculate $L^{cdd}$, we set $\gamma$ to $1.5$. To calculate $L^{icm}$ and $L^{n-icm}$, we initialize $\tau$ to $e^{-1}$. To calculate $L^{joint}$ in the second training stage, we set $\delta$ to $0.1$. For the optimization in each training stage, we apply an AdamW optimizer \cite{loshchilov2017decoupled} with an initial learning rate of $0.0001$ and a weight decay factor of $0.01$. We perform the optimization on $8$ NVIDIA V100 $16$GB GPUs in parallel, and set the mini-batch size on each GPU to $32$. We calculate the average loss on the validation subset after every $10$ epochs. If the resulting number is reduced, then we save the current CDD-ICM model, otherwise we restore the CDD-ICM model to the previous saved version. We decay the learning rate by $50\%$ after each restoration, and terminate the optimization after the $5$th restoration.

\subsection{Experimental Results}
On each dataset, we train a CDD-ICM model by using the training set for optimization and using the validation set for model selection. To compare CDD-ICM with the existing M-CIG methods, we use the test set for evaluation, and finally report the precision, recall, F1 score, and RSIM over all the M-CIG samples in the test set. As shown in Table~\ref{t1}, we achieve SOTA performance on both datasets. Specifically, on CoDraw, CDD-ICM outperforms the existing M-CIG methods by a large margin, where the advantage in precision, recall, and F1 score is relatively larger than that in RSIM. On i-CLEVR, although the existing M-CIG methods did not leave too much room for improvement in precision, recall, and F1 score, CDD-ICM is still better than them in these metrics, reaching almost perfect numbers, and also outperforms them in RSIM.

From all the M-CIG methods in Table~\ref{t1}, we observe two regularities. On the one hand, the performance of these methods on CoDraw is generally worse than that on i-CLEVR. By comparing CoDraw with i-CLEVR, we speculate that this is mainly because the modification texts in CoDraw are commonly longer and more complicated than those in i-CLEVR, which makes it more difficult to generate the desired target images. On the other hand, the performance of these methods in object presence, which is reflected by precision, recall, and F1 score, is generally better than that in object topology, which is reflected by RSIM. By investigating both datasets, we speculate that this is mainly because good performance in object presence just requires correctly identifying the names and attributes of objects from the modification texts, while that in object topology usually requires comprehensively understanding the modification texts.

\subsection{Case Study}
\begin{figure*}[ht!]
\centering
\includegraphics[width=0.98\textwidth,trim={0cm 4.5cm 1.5cm 0cm},clip]{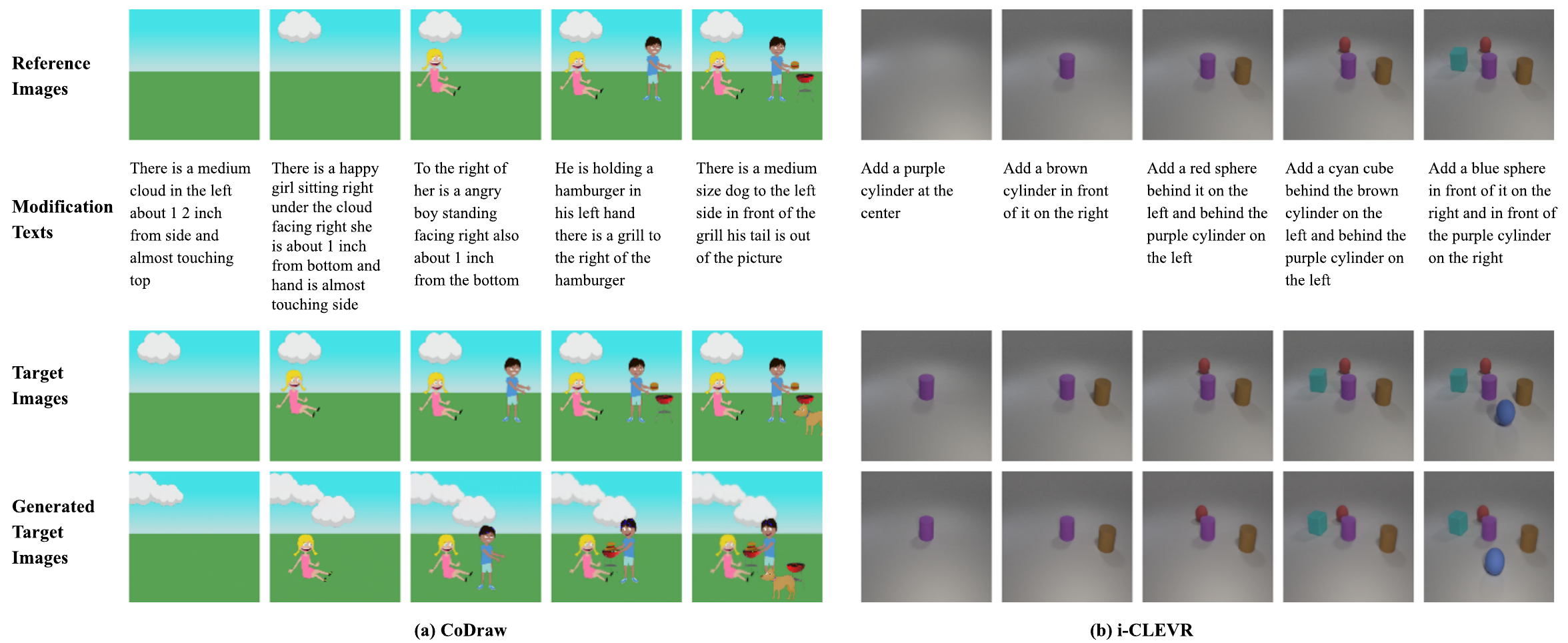}
\caption{Demo cases from CoDraw and i-CLEVR. For the convenience of display, we only include the utterances of the drawer in the modification texts of CoDraw.}
\label{f3}
\end{figure*}
To visually demonstrate the capability of CDD-ICM, we select several representative M-CIG samples from the test set of both datasets, and use the inference results of CDD-ICM on them as demo cases. A demo case from CoDraw and another one from i-CLEVR are shown in Figure~\ref{f3}, and more demo cases are available in the appendix. In the demo cases from CoDraw, the generated target images contain most of the desired objects, but there are still some extra and missing objects. Besides, it also shows that on CoDraw, CDD-ICM struggles with accurately positioning and orienting objects. In the demo cases from i-CLEVR, the generated target images are very similar to the ground-truth target images, where only a few objects are misaligned.

\subsection{Ablation Study}
\begin{table}[ht!]
\small
\centering
\begin{tabular}{@{}ccccc@{}}
\toprule
\multirow{2}{*}{\textbf{Method}} & \multicolumn{2}{c}{\textbf{CoDraw}} & \multicolumn{2}{c}{\textbf{i-CLEVR}} \\
\cmidrule(l){2-3} \cmidrule(l){4-5} & \textbf{F1} & \textbf{RSIM} & \textbf{F1} & \textbf{RSIM} \\
\midrule
CDD-ICM & 89.05 & 57.39 & 99.96 & 85.66 \\
\midrule
w/o ICM & 75.86 & 50.58 & 89.92 & 74.49 \\
\midrule
w/o ICM Guidance & 87.69 & 56.94 & 96.27 & 81.32 \\
\midrule
w/o Classifier-free \\ Guidance & 86.34 & 56.11 & 94.91 & 80.28 \\
\midrule
Fine-tuning of CLIP: \\ Frozen & 80.63 & 54.89 & 87.24 & 72.44 \\
\midrule
Fine-tuning of CLIP: \\ Fully-Trainable & 58.52 & 39.97 & 65.83 & 44.76 \\
\midrule
w/o Iterative Setting & 92.51 & 75.93 & 100.00 & 96.20 \\
\bottomrule
\end{tabular}
\caption{Results of ablation experiments.}
\label{t2}
\end{table}
To probe the performance contribution from each design point of CDD-ICM, we conduct the following five ablation experiments. As shown in Table~\ref{t2}, in each ablation experiment, we change a corresponding design point, and report the resulting F1 score and RSIM on each dataset.

\textbf{ICM.} We disable ICM by skipping the first training stage and setting $\delta$ to $0$ when calculating $L^{joint}$ in the second training stage. As a result, we observe a significant drop in F1 score and RSIM. This verifies the effectiveness of learning ICM as an auxiliary objective of CDD. Besides, we also observe a significant drop in the converging speed of the second training stage. This further verifies that learning ICM is beneficial for learning CDD.

\textbf{ICM Guidance.} We disable ICM guidance by skipping the third training stage and setting $\psi$ to $0$. As a result, we observe a drop in F1 score and RSIM. This verifies the effectiveness of ICM guidance.

\textbf{Classifier-free Guidance.} We disable classifier-free guidance by setting $\lambda$ to $0$ and setting $\phi$ to $1$. As a result, we observe a drop in F1 score and RSIM. This verifies the effectiveness of classifier-free guidance.

\textbf{Fine-tuning of CLIP.} For the fine-tuning of CLIP, which is controlled by the backbone activity ratio $\eta$, we examine two extreme cases. On the one hand, we make CLIP frozen by setting $\eta$ to $0$. On the other hand, we make CLIP fully-trainable by setting $\eta$ to $1$. As a result, we observe a significant drop in F1 score and RSIM in both cases. This verifies the necessity of applying $\eta$.

\textbf{Iterative Setting.} In the evaluation, we disable the iterative setting by using the ground-truth target image at each turn of M-CIG as the reference image of the next turn, which actually downgrades M-CIG to CIG. As a result, we observe a significant rise in F1 score and RSIM. This verifies that M-CIG is more challenging than CIG.

\section{Conclusion and Limitation}
In this paper, we focus on M-CIG, which is a challenging and practical image generation task, and propose a diffusion-based method named CDD-ICM, which achieves SOTA performance on CoDraw and i-CLEVR. The limitation of CDD-ICM mainly lies in its inference efficiency. Although we have accelerated the inference of CDD-ICM by traversing only a part of the time steps in a deterministic manner, it still takes $3$ GPU seconds for CDD-ICM to generate a target image, which is much slower than the GAN-based methods. In the future, we plan to further accelerate the inference of CDD-ICM by applying latent diffusion models \cite{rombach2022high} and knowledge distillation methods \cite{luhman2021knowledge,salimansprogressive}.

{
\small
\bibliographystyle{ieee_fullname}
\bibliography{egbib}
}

\input{appendix}

\end{document}

%% file: appendix.tex
\appendix
\onecolumn

\section{Demo Cases from CoDraw}
\begin{figure}[ht!]
\centering
\includegraphics[width=0.8\textwidth]{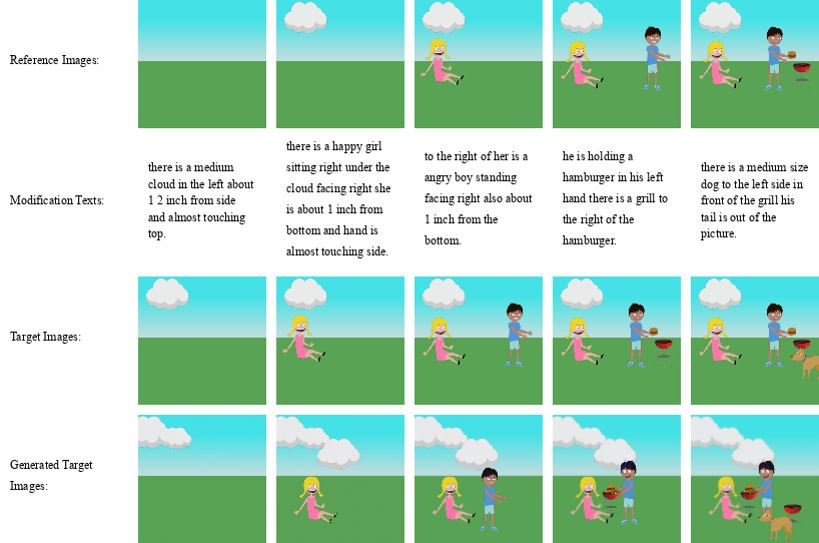}
\caption{}
\end{figure}

\begin{figure}[ht!]
\centering
\includegraphics[width=0.8\textwidth]{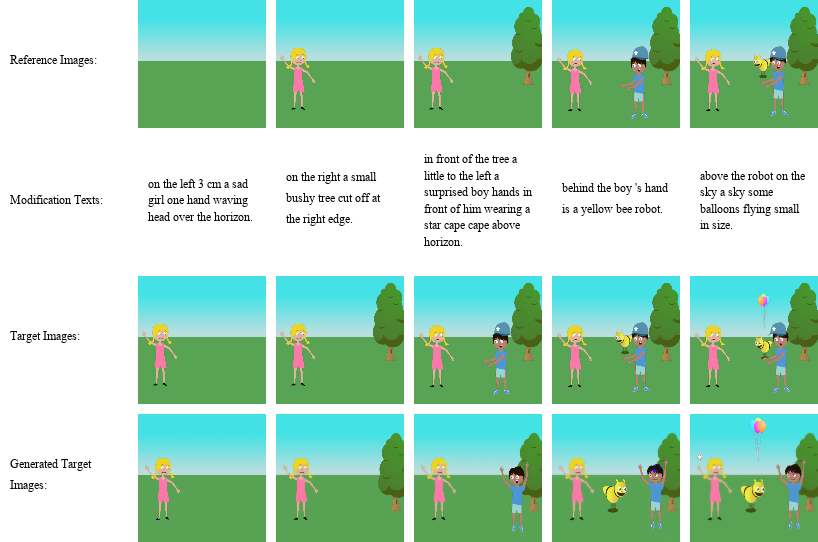}
\caption{}
\end{figure}

\begin{figure}[ht!]
\centering
\includegraphics[width=0.8\textwidth]{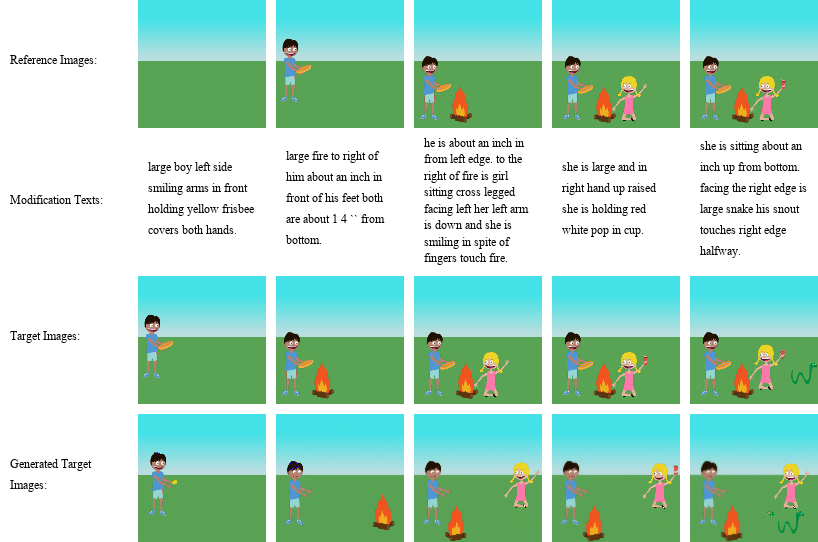}
\caption{}
\end{figure}

\begin{figure}[ht!]
\centering
\includegraphics[width=0.8\textwidth]{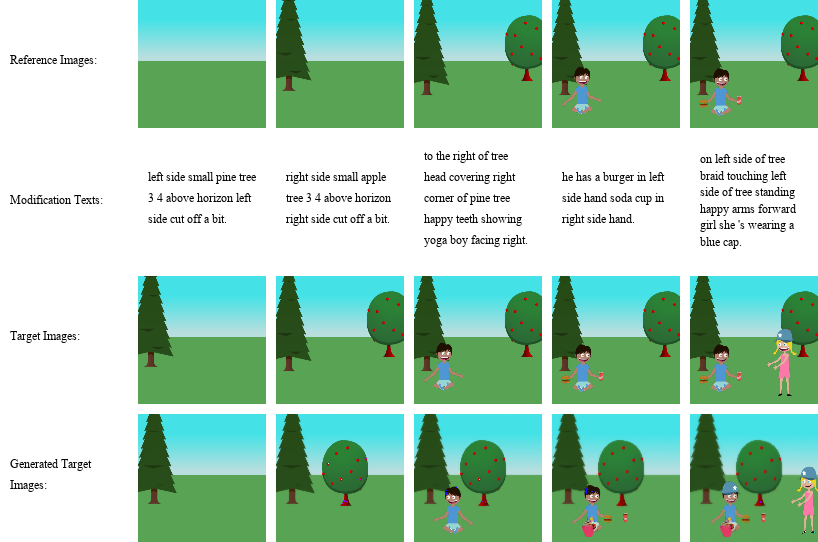}
\caption{}
\end{figure}

\clearpage

\section{Demo Cases from i-CLEVR}
\begin{figure}[ht!]
\centering
\includegraphics[width=0.8\textwidth]{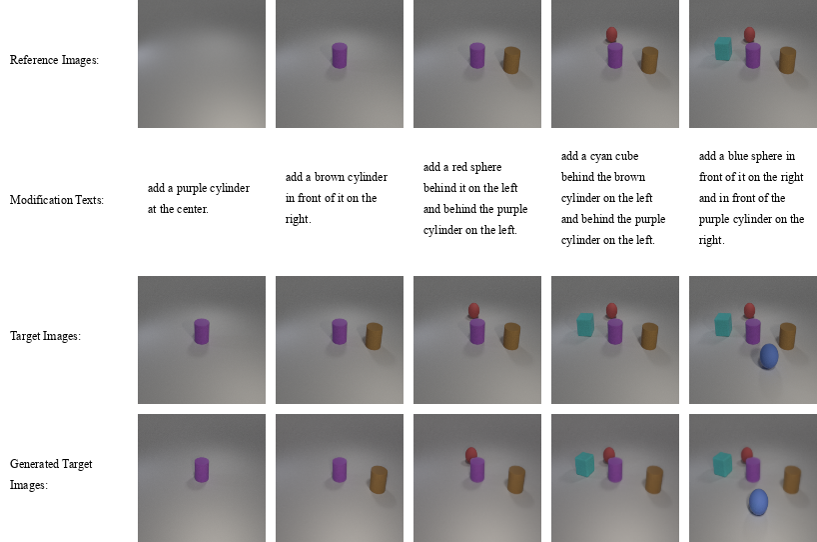}
\caption{}
\end{figure}

\begin{figure}[ht!]
\centering
\includegraphics[width=0.8\textwidth]{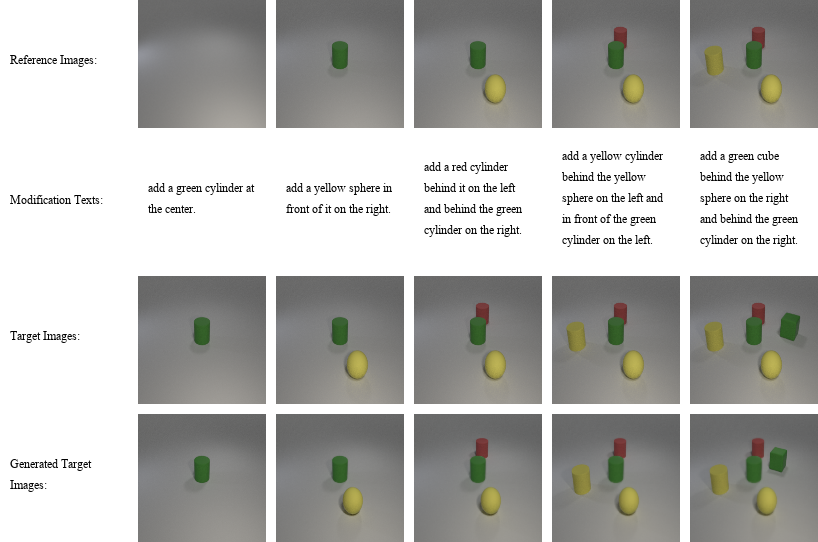}
\caption{}
\end{figure}

\begin{figure}[ht!]
\centering
\includegraphics[width=0.8\textwidth]{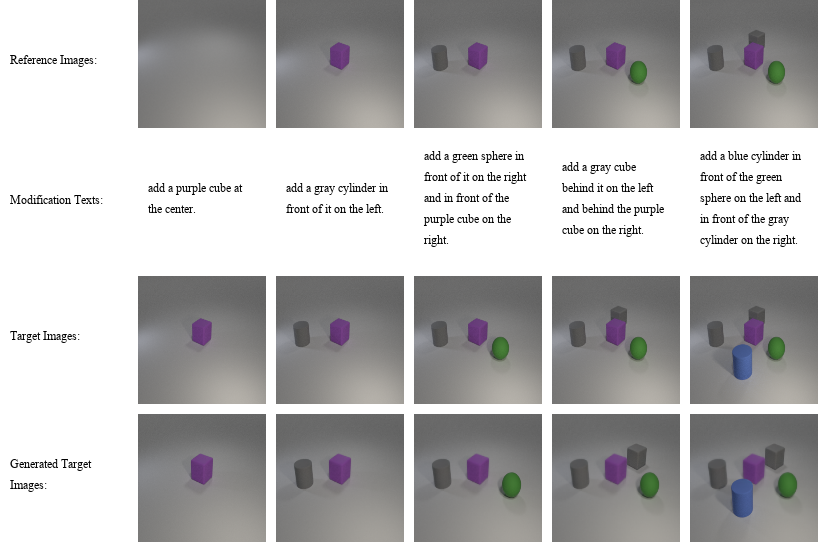}
\caption{}
\end{figure}

\begin{figure}[ht!]
\centering
\includegraphics[width=0.8\textwidth]{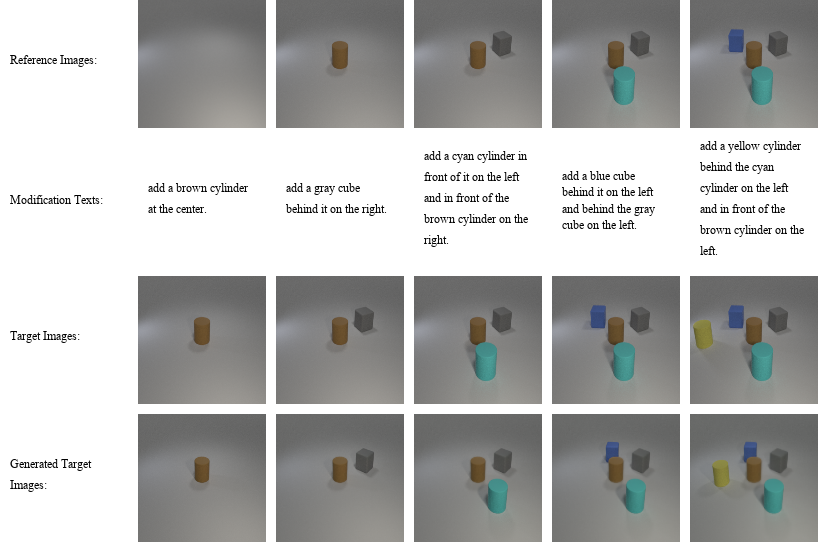}
\caption{}
\end{figure}

%% file: PaperForReview.bbl
\begin{thebibliography}{10}\itemsep=-1pt

\bibitem{Ak2020LearningCR}
Kenan~E. Ak, Ying Sun, and Joo-Hwee Lim.
\newblock Learning cross-modal representations for language-based image
  manipulation.
\newblock {\em 2020 IEEE International Conference on Image Processing (ICIP)},
  pages 1601--1605, 2020.

\bibitem{arjovsky2017towards}
Martin Arjovsky and Leon Bottou.
\newblock Towards principled methods for training generative adversarial
  networks.
\newblock In {\em ICLR}, 2017.

\bibitem{bengio2014deep}
Yoshua Bengio, Eric Laufer, Guillaume Alain, and Jason Yosinski.
\newblock Deep generative stochastic networks trainable by backprop.
\newblock In {\em ICML}, pages 226--234. PMLR, 2014.

\bibitem{brock2019large}
Andrew Brock, Jeff Donahue, and Karen Simonyan.
\newblock Large scale gan training for high fidelity natural image synthesis.
\newblock In {\em ICLR}, 2019.

\bibitem{brock2018large}
Andrew Brock, Jeff Donahue, and Karen Simonyan.
\newblock Large scale gan training for high fidelity natural image synthesis.
\newblock {\em ICLR}, 2019.

\bibitem{brock2017neural}
Andrew Brock, Theodore Lim, James~M Ritchie, and Nick Weston.
\newblock Neural photo editing with introspective adversarial networks.
\newblock In {\em ICLR}, 2017.

\bibitem{dahl2017pixel}
Ryan Dahl, Mohammad Norouzi, and Jonathon Shlens.
\newblock Pixel recursive super resolution.
\newblock In {\em ICCV}, pages 5439--5448, 2017.

\bibitem{dhariwal2021diffusion}
Prafulla Dhariwal and Alexander Nichol.
\newblock Diffusion models beat gans on image synthesis.
\newblock In {\em NeurIPS}, 2021.

\bibitem{dong2014learning}
Chao Dong, Chen~Change Loy, Kaiming He, and Xiaoou Tang.
\newblock Learning a deep convolutional network for image super-resolution.
\newblock In {\em ECCV}, pages 184--199. Springer, 2014.

\bibitem{dong2015image}
Chao Dong, Chen~Change Loy, Kaiming He, and Xiaoou Tang.
\newblock Image super-resolution using deep convolutional networks.
\newblock {\em IEEE transactions on pattern analysis and machine intelligence},
  38(2):295--307, 2015.

\bibitem{dosovitskiy2021an}
Alexey Dosovitskiy, Lucas Beyer, Alexander Kolesnikov, Dirk Weissenborn,
  Xiaohua Zhai, Thomas Unterthiner, Mostafa Dehghani, Matthias Minderer, Georg
  Heigold, Sylvain Gelly, et~al.
\newblock An image is worth 16x16 words: Transformers for image recognition at
  scale.
\newblock In {\em ICLR}, 2021.

\bibitem{el2019tell}
Alaaeldin El-Nouby, Shikhar Sharma, Hannes Schulz, Devon Hjelm, Layla~El Asri,
  Samira~Ebrahimi Kahou, Yoshua Bengio, and Graham~W Taylor.
\newblock Tell, draw, and repeat: Generating and modifying images based on
  continual linguistic instruction.
\newblock In {\em ICCV}, 2019.

\bibitem{fu-etal-2020-sscr}
Tsu-Jui Fu, Xin Wang, Scott Grafton, Miguel Eckstein, and William~Yang Wang.
\newblock {SSCR}: Iterative language-based image editing via self-supervised
  counterfactual reasoning.
\newblock In {\em EMNLP}, 2020.

\bibitem{fu2020sscr}
Tsu-Jui Fu, Xin Wang, Scott Grafton, Miguel Eckstein, and William~Yang Wang.
\newblock Sscr: Iterative language-based image editing via self-supervised
  counterfactual reasoning.
\newblock In {\em EMNLP}, pages 4413--4422, 2020.

\bibitem{goodfellow2014generative}
Ian Goodfellow, Jean Pouget-Abadie, Mehdi Mirza, Bing Xu, David Warde-Farley,
  Sherjil Ozair, Aaron Courville, and Yoshua Bengio.
\newblock Generative adversarial nets.
\newblock In {\em NeurIPS}, 2014.

\bibitem{ho2020denoising}
Jonathan Ho, Ajay Jain, and Pieter Abbeel.
\newblock Denoising diffusion probabilistic models.
\newblock In {\em NeurIPS}, 2020.

\bibitem{ho2022cascaded}
Jonathan Ho, Chitwan Saharia, William Chan, David~J Fleet, Mohammad Norouzi,
  and Tim Salimans.
\newblock Cascaded diffusion models for high fidelity image generation.
\newblock {\em Journal of Machine Learning Research}, 2022.

\bibitem{ho2022classifier}
Jonathan Ho and Tim Salimans.
\newblock Classifier-free diffusion guidance.
\newblock {\em arXiv preprint arXiv:2207.12598}, 2022.

\bibitem{karras2017progressive}
Tero Karras, Timo Aila, Samuli Laine, and Jaakko Lehtinen.
\newblock Progressive growing of gans for improved quality, stability, and
  variation.
\newblock In {\em ICLR}, 2018.

\bibitem{karras2019style}
Tero Karras, Samuli Laine, and Timo Aila.
\newblock A style-based generator architecture for generative adversarial
  networks.
\newblock In {\em CVPR}, pages 4401--4410, 2019.

\bibitem{kenan2020learning}
E~Ak Kenan, Ying Sun, and Joo~Hwee Lim.
\newblock Learning cross-modal representations for language-based image
  manipulation.
\newblock In {\em 2020 IEEE International Conference on Image Processing
  (ICIP)}, 2020.

\bibitem{kim2022diffusionclip}
Gwanghyun Kim, Taesung Kwon, and Jong~Chul Ye.
\newblock Diffusionclip: Text-guided diffusion models for robust image
  manipulation.
\newblock In {\em CVPR}, 2022.

\bibitem{kim2016accurate}
Jiwon Kim, Jung~Kwon Lee, and Kyoung~Mu Lee.
\newblock Accurate image super-resolution using very deep convolutional
  networks.
\newblock In {\em CVPR}, pages 1646--1654, 2016.

\bibitem{kim-etal-2019-codraw}
Jin-Hwa Kim, Nikita Kitaev, Xinlei Chen, Marcus Rohrbach, Byoung-Tak Zhang,
  Yuandong Tian, Dhruv Batra, and Devi Parikh.
\newblock {C}o{D}raw: Collaborative drawing as a testbed for grounded
  goal-driven communication.
\newblock In {\em ACL}, 2019.

\bibitem{kim2019codraw}
Jin-Hwa Kim, Nikita Kitaev, Xinlei Chen, Marcus Rohrbach, Byoung-Tak Zhang,
  Yuandong Tian, Dhruv Batra, and Devi Parikh.
\newblock Codraw: Collaborative drawing as a testbed for grounded goal-driven
  communication.
\newblock In {\em ACL}, pages 6495--6513, 2019.

\bibitem{ledig2017photo}
Christian Ledig, Lucas Theis, Ferenc Husz{\'a}r, Jose Caballero, Andrew
  Cunningham, Alejandro Acosta, Andrew Aitken, Alykhan Tejani, Johannes Totz,
  Zehan Wang, et~al.
\newblock Photo-realistic single image super-resolution using a generative
  adversarial network.
\newblock In {\em CVPR}, pages 4681--4690, 2017.

\bibitem{levy2018generalizing}
Daniel Levy, Matthew~D. Hoffman, and Jascha Sohl{-}Dickstein.
\newblock Generalizing hamiltonian monte carlo with neural networks.
\newblock In {\em ICLR}, 2018.

\bibitem{li2020manigan}
Bowen Li, Xiaojuan Qi, Thomas Lukasiewicz, and Philip~HS Torr.
\newblock Manigan: Text-guided image manipulation.
\newblock In {\em CVPR}, pages 7880--7889, 2020.

\bibitem{liu2020rethinking}
Hongyu Liu, Bin Jiang, Yibing Song, Wei Huang, and Chao Yang.
\newblock Rethinking image inpainting via a mutual encoder-decoder with feature
  equalizations.
\newblock In {\em ECCV}, pages 725--741. Springer, 2020.

\bibitem{liu2022compositional}
Nan Liu, Shuang Li, Yilun Du, Antonio Torralba, and Joshua~B Tenenbaum.
\newblock Compositional visual generation with composable diffusion models.
\newblock In {\em ECCV}, pages 423--439. Springer, 2022.

\bibitem{loshchilov2017decoupled}
Ilya Loshchilov and Frank Hutter.
\newblock Decoupled weight decay regularization.
\newblock In {\em ICLR}, 2019.

\bibitem{lugmayr2020srflow}
Andreas Lugmayr, Martin Danelljan, Luc Van~Gool, and Radu Timofte.
\newblock Srflow: Learning the super-resolution space with normalizing flow.
\newblock In {\em ECCV}, pages 715--732. Springer, 2020.

\bibitem{luhman2021knowledge}
Eric Luhman and Troy Luhman.
\newblock Knowledge distillation in iterative generative models for improved
  sampling speed.
\newblock {\em arXiv preprint arXiv:2101.02388}, 2021.

\bibitem{matsumori2021lattegan}
Shoya Matsumori, Yuki Abe, Kosuke Shingyouchi, Komei Sugiura, and Michita Imai.
\newblock Lattegan: Visually guided language attention for multi-turn
  text-conditioned image manipulation.
\newblock {\em IEEE Access}, 2021.

\bibitem{menon2020pulse}
Sachit Menon, Alexandru Damian, Shijia Hu, Nikhil Ravi, and Cynthia Rudin.
\newblock Pulse: Self-supervised photo upsampling via latent space exploration
  of generative models.
\newblock In {\em CVPR}, pages 2437--2445, 2020.

\bibitem{miyato2018spectral}
Takeru Miyato, Toshiki Kataoka, Masanori Koyama, and Yuichi Yoshida.
\newblock Spectral normalization for generative adversarial networks.
\newblock In {\em ICLR}, 2018.

\bibitem{nam2018text}
Seonghyeon Nam, Yunji Kim, and Seon~Joo Kim.
\newblock Text-adaptive generative adversarial networks: manipulating images
  with natural language.
\newblock {\em NeurIPS}, 2018.

\bibitem{nazeri2019edgeconnect}
Kamyar Nazeri, Eric Ng, Tony Joseph, Faisal~Z Qureshi, and Mehran Ebrahimi.
\newblock Edgeconnect: Generative image inpainting with adversarial edge
  learning.
\newblock {\em arXiv preprint arXiv:1901.00212}, 2019.

\bibitem{nichol2022glide}
Alex Nichol, Prafulla Dhariwal, Aditya Ramesh, Pranav Shyam, Pamela Mishkin,
  Bob McGrew, Ilya Sutskever, and Mark Chen.
\newblock Glide: Towards photorealistic image generation and editing with
  text-guided diffusion models.
\newblock In {\em ICML}, 2022.

\bibitem{nichol2021improved}
Alexander~Quinn Nichol and Prafulla Dhariwal.
\newblock Improved denoising diffusion probabilistic models.
\newblock In {\em ICML}, 2021.

\bibitem{oord2018representation}
Aaron van~den Oord, Yazhe Li, and Oriol Vinyals.
\newblock Representation learning with contrastive predictive coding.
\newblock {\em arXiv preprint arXiv:1807.03748}, 2018.

\bibitem{parmar2018image}
Niki Parmar, Ashish Vaswani, Jakob Uszkoreit, Lukasz Kaiser, Noam Shazeer,
  Alexander Ku, and Dustin Tran.
\newblock Image transformer.
\newblock In {\em ICML}, pages 4055--4064. PMLR, 2018.

\bibitem{paszke2019pytorch}
Adam Paszke, Sam Gross, Francisco Massa, Adam Lerer, James Bradbury, Gregory
  Chanan, Trevor Killeen, Zeming Lin, Natalia Gimelshein, Luca Antiga, et~al.
\newblock Pytorch: An imperative style, high-performance deep learning library.
\newblock In {\em NeurIPS}, 2019.

\bibitem{pathak2016context}
Deepak Pathak, Philipp Krahenbuhl, Jeff Donahue, Trevor Darrell, and Alexei~A
  Efros.
\newblock Context encoders: Feature learning by inpainting.
\newblock In {\em CVPR}, pages 2536--2544, 2016.

\bibitem{radford2021learning}
Alec Radford, Jong~Wook Kim, Chris Hallacy, Aditya Ramesh, Gabriel Goh,
  Sandhini Agarwal, Girish Sastry, Amanda Askell, Pamela Mishkin, Jack Clark,
  et~al.
\newblock Learning transferable visual models from natural language
  supervision.
\newblock In {\em ICML}, 2021.

\bibitem{radford2018improving}
Alec Radford, Karthik Narasimhan, Tim Salimans, and Ilya Sutskever.
\newblock Improving language understanding by generative pre-training.
\newblock {\em OpenAI}, 2018.

\bibitem{ramesh2022hierarchical}
Aditya Ramesh, Prafulla Dhariwal, Alex Nichol, Casey Chu, and Mark Chen.
\newblock Hierarchical text-conditional image generation with clip latents.
\newblock {\em arXiv preprint arXiv:2204.06125}, 2022.

\bibitem{ren2019structureflow}
Yurui Ren, Xiaoming Yu, Ruonan Zhang, Thomas~H Li, Shan Liu, and Ge Li.
\newblock Structureflow: Image inpainting via structure-aware appearance flow.
\newblock In {\em ICCV}, pages 181--190, 2019.

\bibitem{rombach2022high}
Robin Rombach, Andreas Blattmann, Dominik Lorenz, Patrick Esser, and Bj{\"o}rn
  Ommer.
\newblock High-resolution image synthesis with latent diffusion models.
\newblock In {\em CVPR}, 2022.

\bibitem{ronneberger2015u}
Olaf Ronneberger, Philipp Fischer, and Thomas Brox.
\newblock U-net: Convolutional networks for biomedical image segmentation.
\newblock In {\em Medical Image Computing and Computer-Assisted
  Intervention--MICCAI 2015: 18th International Conference, Munich, Germany,
  October 5-9, 2015, Proceedings, Part III 18}, pages 234--241. Springer, 2015.

\bibitem{saharia2022palette}
Chitwan Saharia, William Chan, Huiwen Chang, Chris Lee, Jonathan Ho, Tim
  Salimans, David Fleet, and Mohammad Norouzi.
\newblock Palette: Image-to-image diffusion models.
\newblock In {\em ACM SIGGRAPH 2022 Conference Proceedings}, 2022.

\bibitem{saharia2022photorealistic}
Chitwan Saharia, William Chan, Saurabh Saxena, Lala Li, Jay Whang, Emily
  Denton, Seyed Kamyar~Seyed Ghasemipour, Burcu~Karagol Ayan, S~Sara Mahdavi,
  Rapha~Gontijo Lopes, et~al.
\newblock Photorealistic text-to-image diffusion models with deep language
  understanding.
\newblock {\em arXiv preprint arXiv:2205.11487}, 2022.

\bibitem{saharia2022image}
Chitwan Saharia, Jonathan Ho, William Chan, Tim Salimans, David~J Fleet, and
  Mohammad Norouzi.
\newblock Image super-resolution via iterative refinement.
\newblock {\em IEEE Transactions on Pattern Analysis and Machine Intelligence},
  2022.

\bibitem{salimans2016improved}
Tim Salimans, Ian Goodfellow, Wojciech Zaremba, Vicki Cheung, Alec Radford, and
  Xi Chen.
\newblock Improved techniques for training gans.
\newblock In {\em NeurIPS}, 2016.

\bibitem{salimansprogressive}
Tim Salimans and Jonathan Ho.
\newblock Progressive distillation for fast sampling of diffusion models.
\newblock In {\em ICLR}, 2022.

\bibitem{salimans2015markov}
Tim Salimans, Diederik Kingma, and Max Welling.
\newblock Markov chain monte carlo and variational inference: Bridging the gap.
\newblock In {\em ICML}, pages 1218--1226. PMLR, 2015.

\bibitem{shmelkov2018good}
Konstantin Shmelkov, Cordelia Schmid, and Karteek Alahari.
\newblock How good is my gan?
\newblock In {\em ECCV}, pages 213--229, 2018.

\bibitem{sohl2015deep}
Jascha Sohl-Dickstein, Eric Weiss, Niru Maheswaranathan, and Surya Ganguli.
\newblock Deep unsupervised learning using nonequilibrium thermodynamics.
\newblock In {\em ICML}, 2015.

\bibitem{song2021denoising}
Jiaming Song, Chenlin Meng, and Stefano Ermon.
\newblock Denoising diffusion implicit models.
\newblock In {\em ICLR}, 2021.

\bibitem{song2020denoising}
Jiaming Song, Chenlin Meng, and Stefano Ermon.
\newblock Denoising diffusion implicit models.
\newblock In {\em ICLR}, 2021.

\bibitem{song2017nice}
Jiaming Song, Shengjia Zhao, and Stefano Ermon.
\newblock A-nice-mc: Adversarial training for mcmc.
\newblock {\em NeurIPS}, 30, 2017.

\bibitem{song2019generative}
Yang Song and Stefano Ermon.
\newblock Generative modeling by estimating gradients of the data distribution.
\newblock In {\em NeurIPS}, 2019.

\bibitem{song2021scorebased}
Yang Song, Jascha Sohl-Dickstein, Diederik~P Kingma, Abhishek Kumar, Stefano
  Ermon, and Ben Poole.
\newblock Score-based generative modeling through stochastic differential
  equations.
\newblock In {\em ICLR}, 2021.

\bibitem{song2018spg}
Yuhang Song, Chao Yang, Yeji Shen, Peng Wang, Qin Huang, and C.{-}C.~Jay Kuo.
\newblock Spg-net: Segmentation prediction and guidance network for image
  inpainting.
\newblock In {\em British Machine Vision Conference 2018, {BMVC} 2018,
  Newcastle, UK, September 3-6, 2018}, page~97. {BMVA} Press, 2018.

\bibitem{vo2019composing}
Nam Vo, Lu Jiang, Chen Sun, Kevin Murphy, Li-Jia Li, Li Fei-Fei, and James
  Hays.
\newblock Composing text and image for image retrieval-an empirical odyssey.
\newblock In {\em CVPR}, pages 6439--6448, 2019.

\bibitem{wang2021cycle}
Hao Wang, Guosheng Lin, Steven~CH Hoi, and Chunyan Miao.
\newblock Cycle-consistent inverse gan for text-to-image synthesis.
\newblock In {\em Proceedings of the 29th ACM International Conference on
  Multimedia}, pages 630--638, 2021.

\bibitem{wang-etal-2018-multi-granularity}
Wei Wang, Ming Yan, and Chen Wu.
\newblock Multi-granularity hierarchical attention fusion networks for reading
  comprehension and question answering.
\newblock In {\em ACL}, 2018.

\bibitem{wang2018esrgan}
Xintao Wang, Ke Yu, Shixiang Wu, Jinjin Gu, Yihao Liu, Chao Dong, Yu Qiao, and
  Chen Change~Loy.
\newblock Esrgan: Enhanced super-resolution generative adversarial networks.
\newblock In {\em Proceedings of the European conference on computer vision
  (ECCV) workshops}, pages 0--0, 2018.

\bibitem{wang2018image}
Yi Wang, Xin Tao, Xiaojuan Qi, Xiaoyong Shen, and Jiaya Jia.
\newblock Image inpainting via generative multi-column convolutional neural
  networks.
\newblock {\em NeurIPS}, 2018.

\bibitem{wang2015deep}
Zhaowen Wang, Ding Liu, Jianchao Yang, Wei Han, and Thomas Huang.
\newblock Deep networks for image super-resolution with sparse prior.
\newblock In {\em ICCV}, pages 370--378, 2015.

\bibitem{Wolf2019HuggingFacesTS}
Thomas Wolf, Lysandre Debut, Victor Sanh, Julien Chaumond, Clement Delangue,
  Anthony Moi, Pierric Cistac, Tim Rault, Rémi Louf, Morgan Funtowicz, Joe
  Davison, Sam Shleifer, Patrick von Platen, Clara Ma, Yacine Jernite, Julien
  Plu, Canwen Xu, Teven~Le Scao, Sylvain Gugger, Mariama Drame, Quentin Lhoest,
  and Alexander~M. Rush.
\newblock Huggingface's transformers: State-of-the-art natural language
  processing.
\newblock {\em ArXiv}, 2019.

\bibitem{yan2018shift}
Zhaoyi Yan, Xiaoming Li, Mu Li, Wangmeng Zuo, and Shiguang Shan.
\newblock Shift-net: Image inpainting via deep feature rearrangement.
\newblock In {\em ECCV}, pages 1--17, 2018.

\bibitem{zeng2019learning}
Yanhong Zeng, Jianlong Fu, Hongyang Chao, and Baining Guo.
\newblock Learning pyramid-context encoder network for high-quality image
  inpainting.
\newblock In {\em CVPR}, pages 1486--1494, 2019.

\bibitem{zhang2021text}
Tianhao Zhang, Hung-Yu Tseng, Lu Jiang, Weilong Yang, Honglak Lee, and Irfan
  Essa.
\newblock Text as neural operator: Image manipulation by text instruction.
\newblock In {\em Proceedings of the 29th ACM International Conference on
  Multimedia}, pages 1893--1902, 2021.

\bibitem{zhu2017your}
Shizhan Zhu, Raquel Urtasun, Sanja Fidler, Dahua Lin, and Chen Change~Loy.
\newblock Be your own prada: Fashion synthesis with structural coherence.
\newblock In {\em ICCV}, pages 1680--1688, 2017.

\bibitem{zitnick2013bringing}
C~Lawrence Zitnick and Devi Parikh.
\newblock Bringing semantics into focus using visual abstraction.
\newblock In {\em CVPR}, pages 3009--3016, 2013.

\end{thebibliography}
